\documentclass{article}

\usepackage{microtype}
\usepackage{graphicx}
\usepackage{subfigure}
\usepackage{booktabs} 

\usepackage{hyperref}

\usepackage[accepted]{icml2023}

\usepackage{amsmath}
\usepackage{amssymb}
\usepackage{mathtools}
\usepackage{amsthm}

\usepackage[capitalize,noabbrev]{cleveref}

\theoremstyle{plain}

\theoremstyle{definition}

\theoremstyle{remark}

\usepackage[textsize=tiny]{todonotes}

\begin{document}

\twocolumn[
\icmltitle{An Evaluation of Memory Optimization Methods for Training Neural Networks}



\icmlsetsymbol{equal}{*}

\begin{icmlauthorlist}
\icmlauthor{Xiaoxuan Liu}{ucb}
\icmlauthor{Siddharth Jha}{ucb}
\icmlauthor{Alvin Cheung}{ucb}
\end{icmlauthorlist}

\icmlaffiliation{ucb}{Department of Computer Science, University of California, Berkeley}

\icmlcorrespondingauthor{Xiaoxuan Liu}{xiaoxuan\_liu@berkeley.edu}

\icmlkeywords{Machine Learning, ICML}

\vskip 0.3in
]



\printAffiliationsAndNotice{}  

\newcommand{\lily}[1]{\textcolor{brown}{[Lily: #1]}}
\newcommand{\alvin}[1]{\textcolor{blue}{[Alvin: #1]}}
\newcommand{\sid}[1]{\textcolor{violet}{[Sid: #1]}}
\newcommand{\jianfei}[1]{\textcolor{red}{[Jianfei: #1]}}
\newcommand{\tool}{\textsc{Papaya}\space}

\begin{abstract}
As models continue to grow in size, the development of memory optimization methods (MOMs) has emerged as a solution to address the memory bottleneck encountered when training large models. To comprehensively examine the practical value of various MOMs, we have conducted a thorough analysis of existing literature from a systems perspective. 
Our analysis has revealed a notable challenge within the research community: the absence of standardized metrics for effectively evaluating the efficacy of MOMs. The scarcity of informative evaluation metrics hinders the ability of researchers and practitioners to compare and benchmark different approaches reliably. Consequently, drawing definitive conclusions and making informed decisions regarding the selection and application of MOMs becomes a challenging endeavor.
To address the challenge, this paper summarizes the scenarios in which MOMs prove advantageous for model training. We propose the use of distinct evaluation metrics under different scenarios. By employing these metrics, we evaluate the prevailing MOMs and find that their benefits are not universal. We present insights derived from experiments and discuss the circumstances in which they can be advantageous.

\end{abstract}
\section{Introduction}
To improve accuracy and generalization, deep neural networks (DNNs) have rapidly grown in size~\cite{brown2020language, chowdhery2022palm, touvron2023llama}. The further growth of these models is constrained by hardware limitations. In particular, the limited memory on GPUs restricts how large a model we can train. 
In response, researchers have developed various memory optimization methods (MOMs) to reduce the memory footprint of training DNNs. Popular MOMs include gradient checkpointing, swapping, and activation reduced training (ACT). We describe these in detail in \autoref{sec:different-moms}.

In this study, we conduct a survey of MOMs used in DNN training to understand when MOMs prove advantageous. Specifically, we aim to address several key questions: Does the application of these techniques truly improve the training process? Which technique achieves the best time-memory trade-off? Are there MOMs that demonstrate superior performance on particular architectural configurations? 

In short, consistent with previous literature, we find that MOMs help in two ways. First, they reduce the peak memory compared to standard training for a fixed batch size. Second, they enable larger maximum batch sizes compared to standard training for a given hardware configuration. 
However, our central finding is that it is difficult to answer our motivating questions with existing ways of evaluating MOMs.
Most of these issues stem from the absence of standard networks, metrics, and experimental practices. 
Crucially, previous evaluation metrics only consider a single aspect of training and ignore the application scenarios of MOMs. Some of these metrics even conflict with each other.
For example, as shown in \autoref{sec:experiment}, synchronous swapping performs better than gradient checkpointing on the Swin-Large model when evaluated with maximum batch size, but gradient checkpointing is more efficient when considering the execution overhead at a given batch size compared with the original training. 


In this paper, we provide a comprehensive overview of various MOMs and outline their respective application scenarios. Recognizing the diversity of these scenarios, we propose using distinct evaluation metrics tailored to each scenario. By employing these evaluation metrics, we conduct a thorough assessment of numerous mainstream network architectures. Surprisingly, our findings reveal that MOMs only demonstrate benefits in particular settings. 
Additionally, we provide a detailed discussion of the conditions under which applying MOMs is advantageous, thereby aiding future researchers in making informed decisions regarding the selection and adoption of MOMs.

In summary, the paper makes the following contributions: (1) We provide a summary of MOM application scenarios and propose tailored evaluation metrics for each scenario. (2) We conduct a comprehensive evaluation of popular MOMs in widely-used frameworks, analyzing their performance in both single GPU and distributed settings.


\section{Overview of Memory Optimization Methods}
\label{sec:related-work}
\subsection{Memory Consumption in Model Training}
The bulk of memory usage during training falls under three categories: 
(1) \emph{activations}, which are intermediate tensors stored during the forward pass that are needed to compute gradients during the backward pass. 
(2) \emph{model related objects} that store the model's state, such as model parameters (e.g., weight matrices), optimizer states, and gradient accumulators.
(3) \emph{workspace memory}, which includes temporary storage for calculations. Workspace memory will be released once the operator finishes execution and typically consumes only a small fraction of memory.
\subsection{Different MOMs Evaluated in this Paper}
\label{sec:different-moms}
This study mainly focuses on memory optimization methods (MOMs) from a system perspective. Apart from MOMs, other approaches such as low-rank kernels~\cite{katharopoulos2020transformers, peng2021random} and leveraging sparsity~\cite{kitaev2020reformer, daras2020smyrf} can lower memory consumption. These alternative methods fall under the algorithmic perspective, often involving modifications to the kernels or network architecture.
Our current research is primarily focused on exploring system-level memory optimization methods.


\textbf{Gradient Checkpointing}~\cite{chen2016training} trades computation for memory by dropping some of the activations during the forward pass from memory and recomputing them during the backward pass.
\citet{jain2020checkmate, kumar2019efficient} formalize the problem of trading computation for memory as a linear programming problem to generate a near-optimal schedule. To handle not only static but also dynamic computation graphs, \citet{kirisame2020dynamic} proposes a greedy online algorithm for heuristically checkpointing arbitrary models.

\textbf{Activation Compressed Training (ACT)} \cite{pmlr-v162-liu22v, pan2021mesa, liu2022exact} is a technique that compresses the training activation during the forward pass, and then decompresses the saved activation during back propagation to calculate the gradient. 
It has been successfully applied to a variety of models such as convolutional neural networks, transformers, and graph neural networks \cite{pmlr-v162-liu22v, pan2021mesa, liu2022exact}. In exchange for lower memory consumption, ACT increases computation and may negatively impact the model's accuracy due to lossy compression.
 
\textbf{Swapping} Swapping offloads activations to external memory (e.g., CPU memory), which comes at the cost of transferring data to another storage and can increase execution time.
\cite{wang2018superneurons, huang2020swapadvisor, peng2020capuchin} explore the search space of jointly optimizing operator scheduling, memory allocation, and swap decisions for static graphs. 
Swapping can be applied to memory consumption other than activations. ZeRO-Offload~\cite{ren2021zero} offloads gradients and optimizer states to the CPU, but stores parameters and activations on the GPU.
Recent work~\citep{beaumont2021efficient} also explores combining gradient checkpointing and offloading.

\begin{figure}
\centering
\includegraphics[width=\linewidth]{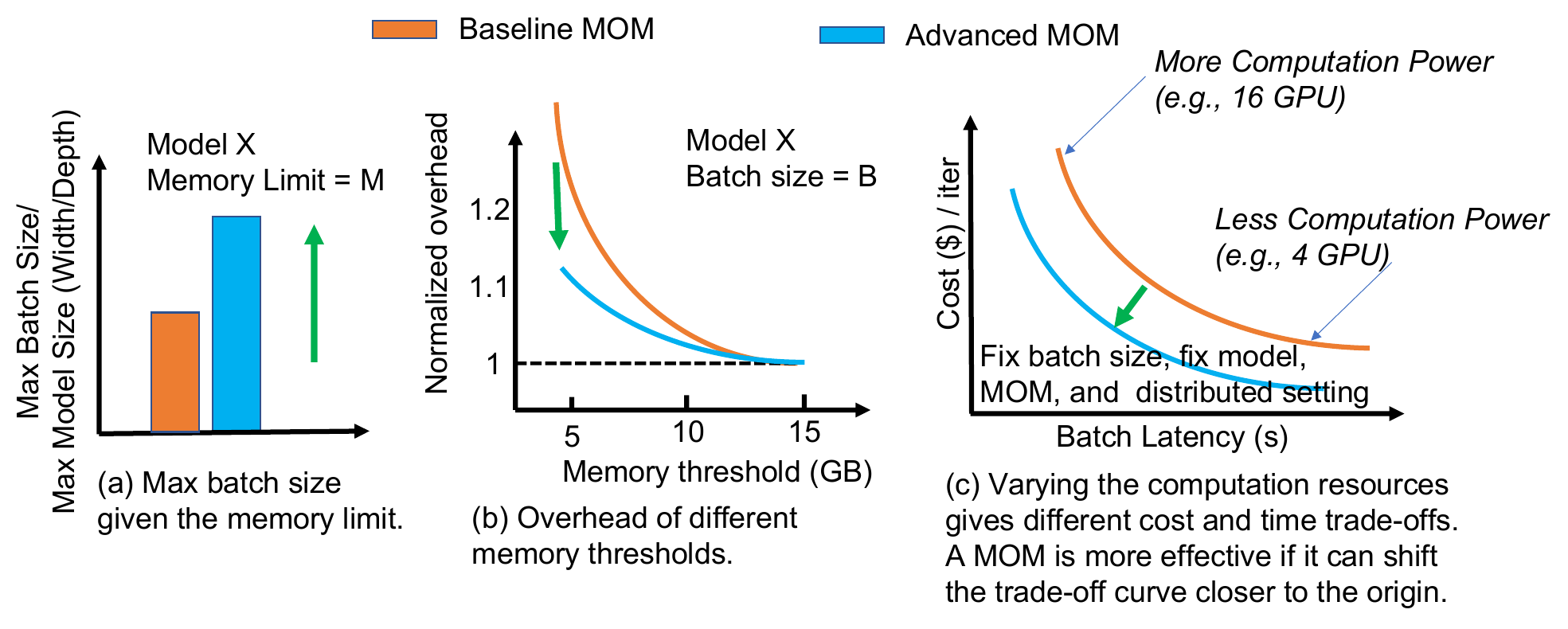}
\caption{Evaluation metrics for MOMs.}
\label{fig:ill_all}
\end{figure}

\begin{figure*}[t]
\centering
\includegraphics[width=0.9\linewidth]{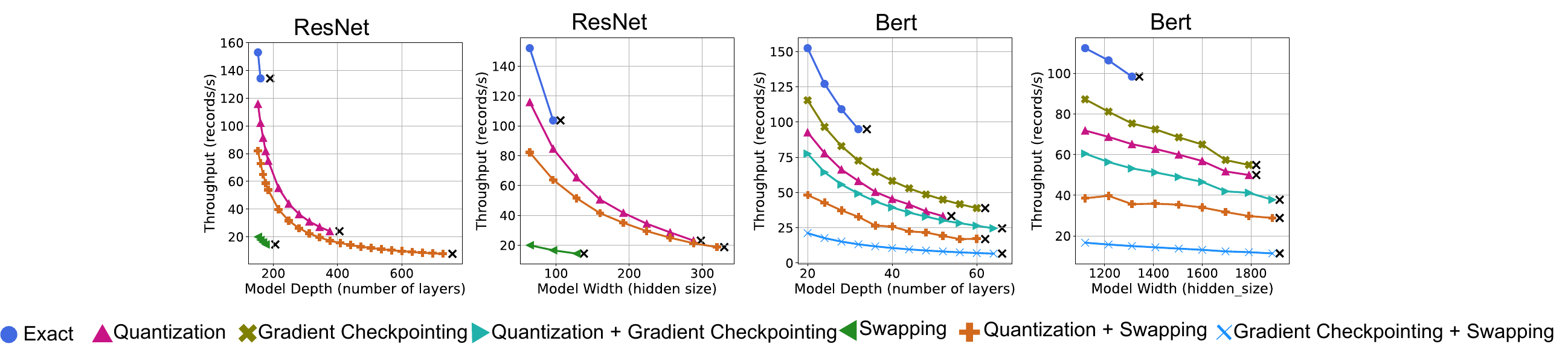}
\caption{Training throughput of different MOMs when varying model depth and width. Batch size is fixed across different training settings. For ResNet, batch size=64. For Bert, batch size=24. The cross indicates an out-of-memory error.}
\label{fig:model_size}
\end{figure*}

\begin{figure*}
\centering
\includegraphics[width=\linewidth]{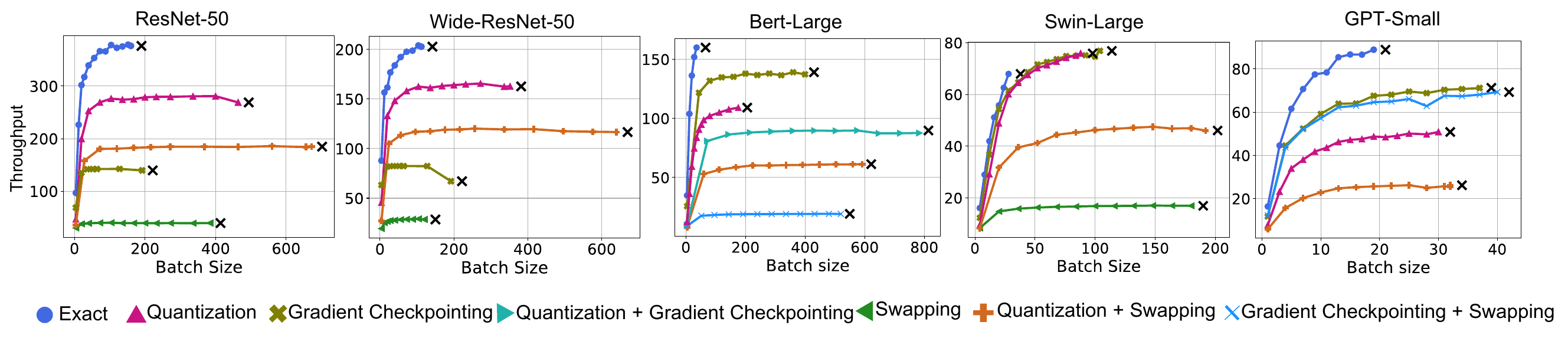}
\caption{Training throughput of different MOMs when varying batch size on a single V100. The cross indicates an out-of-memory error.}
\label{fig:case_study_all}
\end{figure*}

\section{MOM Application Scenario and Evaluation Metrics}

Previous evaluations of various MOMs can be categorized as follows:
(1) \citet{chen2021actnn, peng2020capuchin, pmlr-v162-liu22v} compare the largest batch size the MOM enables given the same memory budget as shown in Figure~\ref{fig:ill_all}(a), which is limited by the memory capacity of the hardware.
(2) Similarly, \citet{chen2016training, chen2021actnn, pmlr-v162-liu22v} compare the largest model the MOM allows given the same batch size and memory budget, using the largest depth and width of the model as a measure of efficacy.
(3) \citet{kirisame2020dynamic, chen2021actnn, huang2020swapadvisor, jain2020checkmate, pmlr-v162-liu22v} compare the performance overhead of a MOM with the original training under different memory thresholds, given the same batch size on a specified model, as shown in Figure~\ref{fig:ill_all} (b).

However, there is an existing gap between these evaluation metrics and practical use. For those wanting faster training, a larger batch size may not result in higher throughput due to the MOM's overhead. For those wanting to train larger models, looking at just the largest trainable model is not sufficient as it must be trained using a realistic throughput, which the MOM hinders. Thus we argue that MOMs should be evaluated in the context of their application. We list three main scenarios in which practitioners may find it useful to employ MOMs.



\textbf{Training larger models: }\textit{Fixed hardware, evaluated with the maximum model size that satisfies a minimum throughput threshold}: The first scenario involves a fixed hardware setting with the goal of training a larger model within the memory limit. 
Under this scenario, evaluating based on the model size is essential. However, it is equally important to consider the training speed. To address this, we propose a new evaluation metric: the maximum model size that satisfies a minimum throughput threshold. This metric takes into account both the model size expansion achieved through MOM and the practical feasibility of training in terms of throughput.

\textbf{Faster training: }\textit{Fixed hardware, evaluated with the maximum throughput}: The second scenario involves a fixed hardware setting and aims to enable a larger batch size, reducing the number of iterations required to train a model while increasing device utilization and speeding up the training process.
In this case, the evaluation metric to be used is the maximum throughput. As demonstrated in \autoref{sec:experiment}, throughput increases with larger batch sizes. Therefore, the maximum throughput is attained when the model is trained using the maximum batch size. This metric provides a measure of the highest achievable training speed, indicating the effectiveness of MOM in improving the overall throughput of the training process.

\textbf{Cheaper training: }\textit{Fixed training setting, evaluated with the cost-latency trade-off curve}: Finally, the third scenario involves a fixed training setting, but the hardware configuration is varied to find the best cost/time trade-off with and without MOM. In this setting, the monetary cost of training a model is directly proportional to the number of GPUs employed and the time taken for training. By leveraging MOM, it is possible to reduce the required number of GPUs, resulting in a potentially lower total training cost, despite an extended training time. Therefore, when the model, batch size, and training setting are held constant, a trade-off frontier is established, as illustrated in \autoref{fig:ill_all} (c). 
The practitioner's preferences (time vs cost) determine what point on the frontier is chosen for training.

\section{How good are existing MOMs, really?}
\label{sec:experiment}

\begin{figure}
\centering
\includegraphics[width=0.96\linewidth]{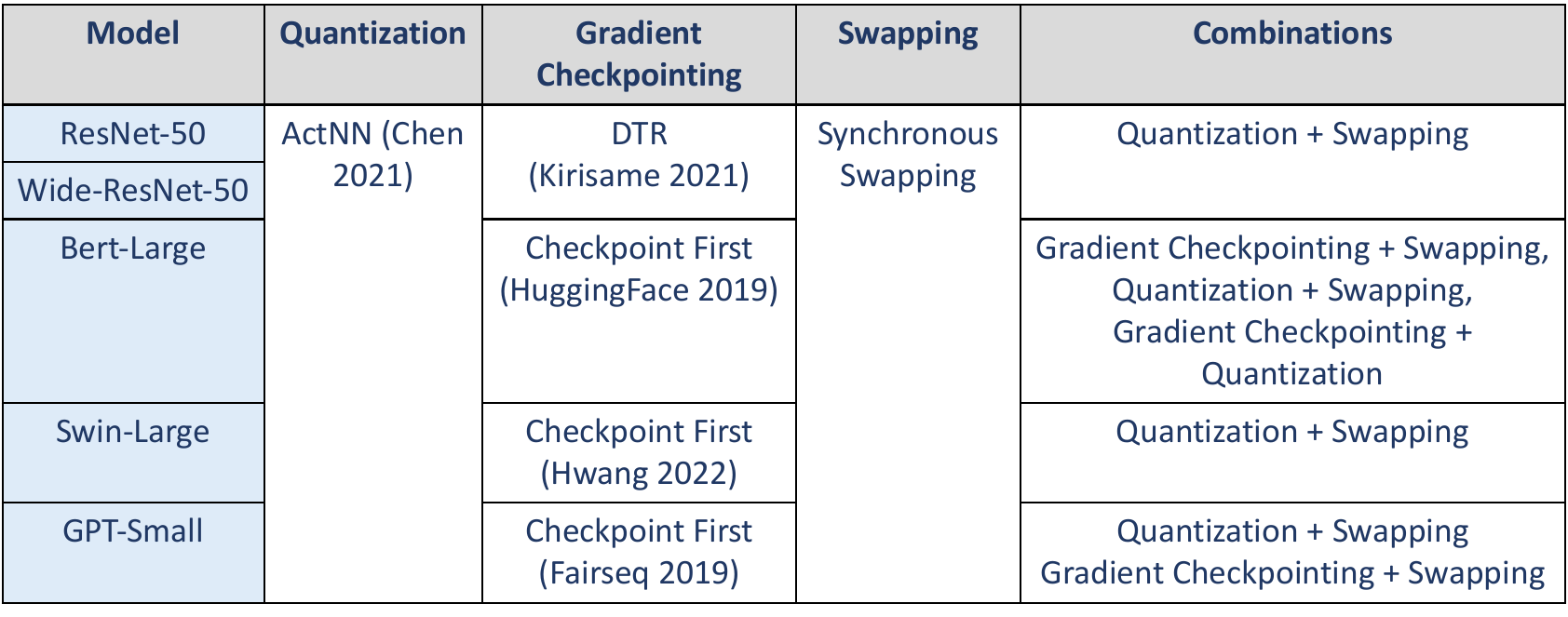}
\caption{MOMs evaluated in this paper.}
\label{fig:case_study_method}
\end{figure}

\begin{figure*}
\centering
\includegraphics[width=0.8\linewidth]{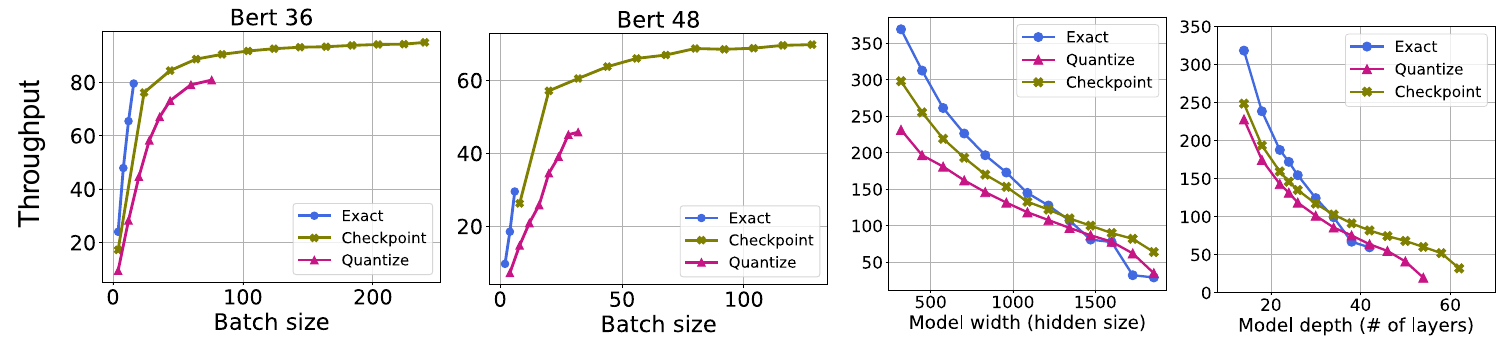}
\caption{Case Study on Bert.}
\label{fig:case_study2}
\end{figure*}

\subsection{Experiment Setup}
\textbf{Evaluated Models.} We evaluate different MOMs on both convolutional neural networks (CNN) and transformer-based models. For CNNs, we evaluate ResNet-50~\cite{he2016deep} and Wide-ResNet-50~\cite{zagoruyko2016wide}. For transformer-based models, we test both language tasks (Bert~\cite{devlin2018bert}, GPT~\cite{radford2021learning}) and vision tasks (Swin-Large~\cite{liu2021swin}).

\textbf{Evaluated MOMs.}  We list all evaluated MOMs and their combinations in \autoref{fig:case_study_method}. For the state-of-art activation compressed training, we utilize the ActNN~\cite{chen2021actnn} quantizer, which quantizes each element of the activation layer into four bits. For gradient checkpointing on CNNs, we choose DTR~\cite{kirisame2020dynamic}, a greedy online algorithm that evicts activations based on different heuristics when the peak memory exceeds the hardware limit. For transformer-based models we use the widely-adopted checkpointing policy that checkpoints the input of each transformer block. We implement synchronous swapping in PyTorch ourselves. Additionally, we test the combinations of different MOMs as shown in \autoref{fig:case_study_method}.

\subsection{Maximum Model Size that Satisfies a Minimum Throughput Threshold}
In this section, we try to answer the question of whether MOMs can enable training larger models without excessive overhead. We investigate the maximum model size that different MOMs can train on a single V100 GPU. By fixing the batch size, we record the training throughput of ResNet/Bert models with varying depths/widths. As depicted in \autoref{fig:model_size}, we observe a decrease in throughput as the model size increases across all training settings, as larger models demand more computation capability. 


\begin{figure*}[t]
\begin{center}
    \includegraphics[width=\linewidth]{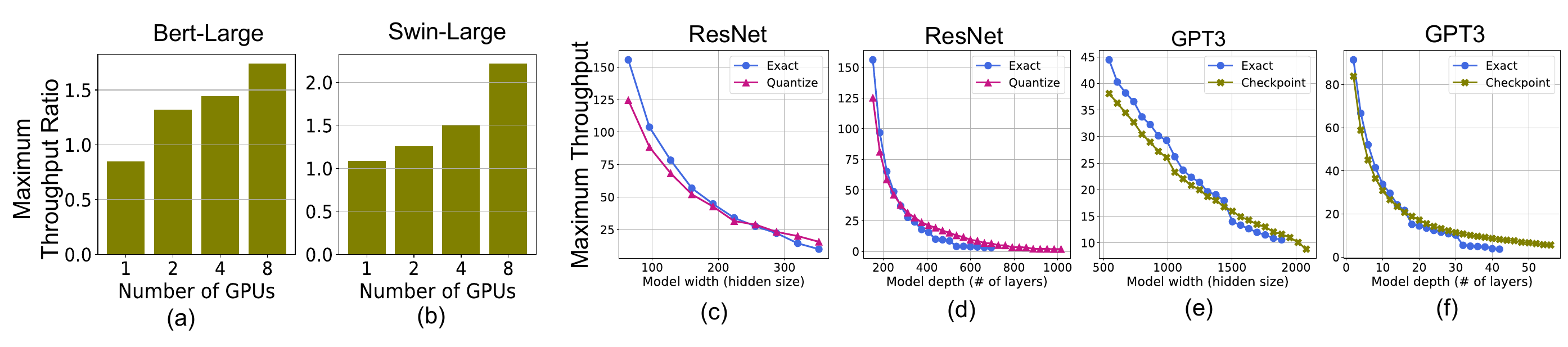}
\end{center}
\caption{(a), (b) show the maximum throughput ratio of training with 1, 2, 4, 8 V100 GPUs on Bert-Large and Swin-Large. (c) - (f) shows the maximum throughput of training ResNet and GPT with different depths and widths.}
\label{fig:impl}
\end{figure*}

\subsection{Maximum Throughput on Various Models}
In this section, we try to determine if MOMs can enable faster training by increasing the maximum throughput.
We test the throughput of different training settings by varying the batch size on a single V100.
As shown in \autoref{fig:case_study_all}, the throughput of most training instances increases with batch size, although to varying degrees. The only exception is DTR on Wide-ResNet-50, where the maximum throughput drops from 80.6 records/s to 65.1 records/s as batch size increases from 160 to 200. This is because DTR determines the rematerialization policy based on the memory limit dynamically, and evicts more activations as batch size increases and memory budget decreases.

All training instances display diminishing returns as batch size increases. For the original training on Bert-Large and Swin-large, the training process is aborted due to out-of-memory error before reaching a throughput plateau, whereas all other training instances reach a plateau where increasing the batch size barely increases training throughout.

Lastly, when compared to the original training, all MOMs increase the maximum batch size. On average, different MOMs can increase the maximum batch size by 2.6$\times$, 2.8$\times$, 11.2$\times$, 4.5$\times$, 1.8$\times$ for the five evaluated models, which aligns with previous work.

However, when considering the purpose of increasing the maximum throughput as mentioned above, only gradient checkpointing and quantization on Swin-Large increase the maximum throughput by 8.8\% (76.86 vs 70.65) and 6.7\% (75.35 vs 70.65) respectively. On average, the maximum training throughput are only 43.6\%, 46.1\%, 50.6\%, 76.6\%, 64.4\% of the original on five evaluated models when MOMs are applied.
In this case, the two metrics (maximum batch size vs. maximum throughput) lead to completely opposite conclusions: MOMs are promising when evaluated using maximum batch size as the metric, while training is actually slowed down when evaluated using maximum throughput, which argues that the majority of MOMs are actually detrimental for faster training.

\subsubsection{Case Study on Bert}
Next, we ask if MOMs can be used to improve {\em both} batch size and throughput. From our results shown in \autoref{fig:case_study_all}, training with quantization and gradient checkpointing are the most promising as they achieve the highest maximum throughput among all MOMs. 
Therefore, we apply these two MOMs on various Bert model sizes to see if they improve the maximum throughput. 
We first conduct experiments on Bert models with 36 layers and 48 layers.
As shown in \autoref{fig:case_study2}, for Bert with 36 layers, the maximum training throughput with gradient checkpointing marginally exceeds that of the original. For the Bert 48 model, both training with quantization and checkpointing improve throughput, improving maximum throughput by 80.1\% (38.7 vs 69.7) and 32.3\% (38.7 vs 51.2) respectively.

\subsection{Cost Time Trade-off}
\begin{figure}
\centering
\includegraphics[width=0.7\linewidth]{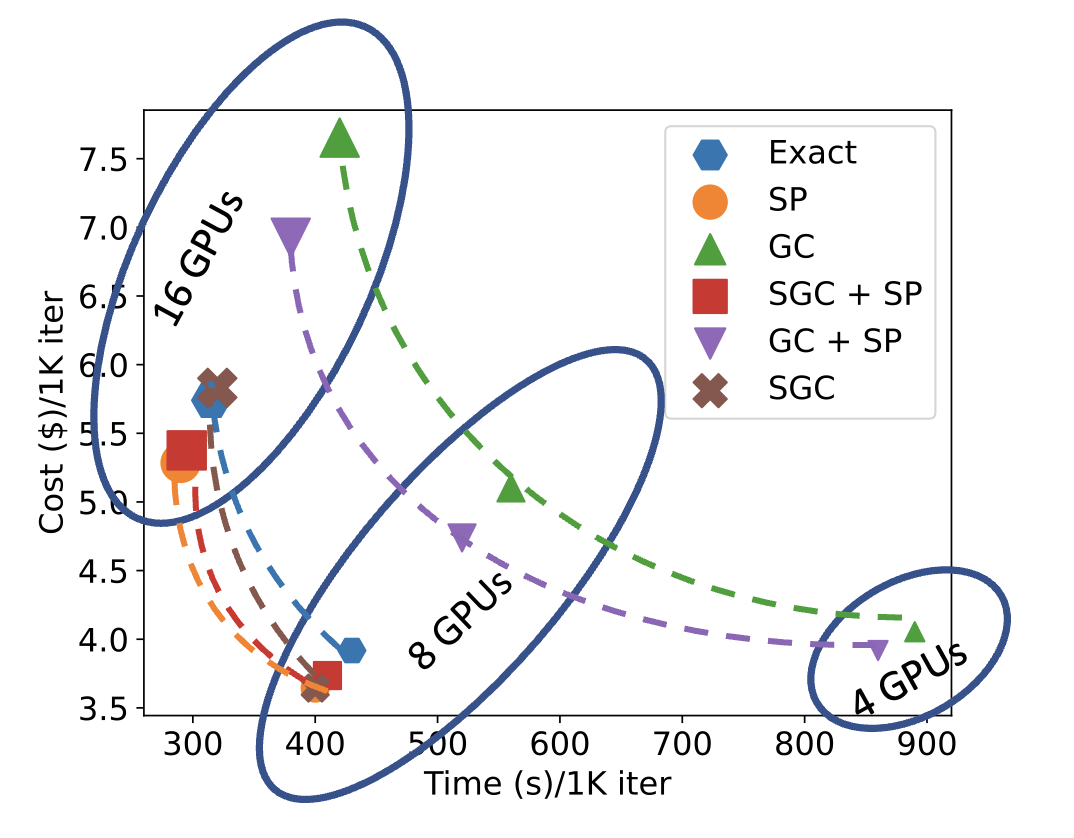}
\caption{Training cost (money cost to run iteration) and batch latency trade-off curve of different MOMs on GPT 6.7B. All training settings use model parallelism. GC: Naive gradient checkpointing. SGC: Selective gradient checkpointing. SP: sequence parallel. Exact: Original training without MOM. We also test methods combination (SGC+SP, GC+SP).}
\label{fig:tradeoff}
\end{figure}
In this experiment, we seek to know if applying MOM can reduce the training cost.
We test the GPT 6.7B model on a single a2-megagpu-16g GCP node~\cite{a2}, with 16 A100 GPUs connected with NVLink, 96 vCPUs, and 1360 GB memory. We tested gradient checkpointing (GC) that only recomputes the input of each transformer block, selective gradient checkpointing (SGC)~\cite{korthikanti2022reducing} that only recomputes the activations within the attention layers, sequence parallel~\cite{korthikanti2022reducing} and their combination provided with Megatron~\cite{megatron}. For all evaluated settings, we apply the same tensor parallelism setting.

As shown in \autoref{fig:tradeoff}, only GC and GC+SP can fit the training on 4 GPUs due to naive gradient checkpointing that only stores the input of each transformer block and greatly reduces stored memory. However, using GC is ineffective in terms of the total cost of training a single iteration, as it has a higher per-iteration cost than not applying any MOM on 8 GPUs.
Although using GC can reduce the total number of nodes, the overall monetary cost remains higher than the original training due to its significant overhead. 

Moreover, we found that using only sequence parallel provides the best latency-cost tradeoff compared to other training settings, as it does not introduce computation or communication overhead when applied with tensor parallelism. It divides the computation across the sequence dimension, reducing the duplicated computation and speeding up the overhead training with the same hardware.


\section{When is MOM more beneficial?}
In this section, we present two key findings regarding the optimal utilization of MOMs for model training. Subsequently, we conduct experiments to validate and substantiate these findings.

\textbf{MOMs are more beneficial for larger models}. In our study, we concentrate on comparing models within the same family (e.g., different layers of Bert models, GPT models with varying numbers of encoder/decoder blocks, etc.), as models from different families are not directly comparable. As models grow larger in size, the likelihood of the original training being underutilized increases. Therefore, the application of MOM becomes more advantageous as it enables an increase in batch size and improves hardware utilization. 

To validate this implication, we conducted experiments by varying the depth and width of ResNet and GPT-3 models on V100 GPUs. The results, as depicted in Figure~\ref{fig:impl}, demonstrate that when the model is shallow enough (with fewer than 280 layers for ResNet or fewer than 18 layers for GPT-3) or narrow (with a hidden size smaller than 248 for ResNet or 1500 for GPT-3), the application of MOMs actually slows down the training process. However, as the model becomes deeper or wider, MOMs start to accelerate the training. In the most extreme case, the original training fails to execute even with a batch size of one, emphasizing the necessity of applying MOMs to enable single-GPU training. These findings provide evidence supporting the claim that the benefits of MOMs are more pronounced with increasing model depth and width.

\textbf{MOMs are more beneficial for multi-GPU data parallel training.} The intuition behind this is that communication makes MOMs less expensive. Communication and synchronization across multiple devices result in a substantially higher fixed execution cost. 

We compare the maximum throughput ratio between gradient checkpointing and the original training on Bert-Large and Swin-Large on distributed data-parallel training with 1, 2, 4, 8 V100 GPUs on a single machine. As shown in Figure~\ref{fig:impl}, the ratio increases by adding more GPUs. Specifically, on Bert-Large, gradient checkpointing slows down training on a single GPU setting. However, as we increase the number of GPUs, gradient checkpointing achieves a higher maximum throughput than the original training, and can speed up the training process by 1.7× when there are 8 GPUs.

\section{Conclusion}
In this study, we present a comprehensive analysis of the benefits of applying various memory optimization methods (MOMs) in different scenarios. By considering distinct application scenarios, we propose to utilize specific evaluation metrics tailored to each scenario. To assess the effectiveness of existing popular MOMs in mainstream frameworks, we conduct experiments using the newly introduced evaluation metrics. The results demonstrate that the application of MOMs does not always yield favorable outcomes. Thus we urge practitioners to exhibit careful judgement and reflect on their use case before applying MOMs.

\nocite{langley00}

\bibliography{example_paper}
\bibliographystyle{icml2023}

\newpage
\appendix
\onecolumn

\end{document}